# Low-Rank Isomap Algorithm

**Eysan Mehrbani · Mohammad Hossein Kahaei[1]**

**Abstract**
The Isomap is a well-known nonlinear dimensionality reduction method that highly suffers from computational complexity. Its computational complexity mainly arises from two stages; a) embedding a full graph on the data in the ambient space, and b) a complete eigenvalue decomposition. Although the reduction of computational complexity of the graph stage has been investigated, yet the eigenvalue decomposition stage remains a bottleneck in the problem. In this paper, we propose the Low-Rank Isomap algorithm by introducing a projection operator on the embedded graph from the ambient space to a low-rank latent space to facilitate applying the partial eigenvalue decomposition. This approach leads to reducing the complexity of Isomap to a linear order while preserving the structural information during the dimensionality reduction process. The superiority of the Low-Rank Isomap algorithm compared to some state-of-art algorithms is experimentally verified on facial image clustering in terms of speed and accuracy.

## 1 Introduction

Dimensionality reduction is an indispensable part of modern data processing tasks due to the massive amount of information provided by pervasive means of sensing and storage. Instances of high-dimensional processing tasks are hyperspectral and medical image processing, data visualization and recognition/classification. Depending on the nature of the problem, the dimensionality of the ambient space may vary. On the other hand, the inherent dimensions of the data within the space tends to remain [1]. Here, dimensionality reduction algorithms aim at finding the most succinct structure of the data embedded in the ambient space [2]. As the intrinsic geometry of the data in the ambient space does not necessarily lie near a linear arrangement, increasing demand for nonlinear algorithms of capturing the low dimensional structures has been arisen [3].

Classical Multi-dimensional Scaling (MDS) [4] is a linear mapping method that has majorly contributed to manifold based algorithms. The MDS finds a low dimensional mapping so that to preserve the pairwise geodesic distance between the original data points. When the data structure grows nonlinear, the geodesic distance will largely deviate from the Euclidean distance. As the geodesic is indeed a metric to be preserved in order to properly capture the data structure, MDS fails when facing nonlinearities in the data.

The methods using the geodesic distance to obtain the underlying manifold are either local or global [5]. The local methods represent each data point as a combination of the other points in the neighborhood [2, 6], whereas the global methods seek the overall geometry of the data as a whole [7]. Isomap is a global method built upon the MDS model by computing the geodesic distance as a preserving metric. The geodesic is calculated by constructing a graph based on the global graph structure of the data. The weight to each edge is then calculated using common shortest path finding algorithms such as the Dijkstra or A* [7]. Ultimately, a linear mapping is performed onto a low-dimensional space so that to preserve the graph edge weights.

---

[1] School of Electrical Engineering, Iran University of Science and Technology, kahaei@iust.ac.ir.

The global Isomap highly suffers from computational complexity due to the time consumption of graph building and geodesic distance estimation, especially when the number of observations grow large. Many attempts have been made to reduce the complexity of this algorithm by specifying a geodesic distance metric exclusively to the data point is one neighborhood. The neighborhood setting for each data point is a connection network to its k-nearest neighbors where the distance to each locally adjacent observation is the shortest path length. In the following section a general overview of the variations over the basic Isomap algorithm is presented., Then, this paper argues about the impracticability of the previous methods with the most accurate outcomes and proposes a novel method with computational complexity of order $N$.

The remnant of this paper is organized as follows. In Section 2 the computational complexity of the Isomap algorithm and the potential problems are studied. In Section 3 the proposed method is studied and in Section 4 numerical results are presented. A. Section 5 concludes the paper.

## 2  Isomap and  Computational Complexity

Formally speaking, for a set of observations $\mathbf{X} = [x_1, x_2, \ldots, x_N]$ with $x_i \in \mathbb{R}^M; i = 1,2,\ldots,N$, the dimensionality reduction problem maps all the data points in the ambient space –which normally is of high dimensionality $M$ - onto a latent space of lower dimensionality $m$ as the solution to the optimization problem of (1).

$$\min_{X' \in \mathbb{R}^m} \sum_{l \neq k} (d(x_l - x_k) - d(x'_l - x'_k))^2 \tag{1}$$

Where $\mathbf{X}' = [x'_1, x'_2, \ldots, x'_N]$ with $x'_i \in \mathbb{R}^m; i = 1,2,\ldots,N$ denote the mapped data points onto the latent space. This optimization problem aims at preserving the locality information of the data structure in the course of projection. In MDS, the locality information is considered as the neighborhoods established upon the Euclidean distance metric [12], whereas Isomap and its variants in discrete applications use graphical distance based on a graph defined as below given a specific definition of the neighborhood.

$$d_G(x_i, x_j) = \begin{cases} d_X(x_i, x_j) & if\ x_i\ and\ x_j\ are\ neighbors \\ \infty & otherwise \end{cases}$$

As the number of data points increases, the graphical distance will be a better approximation of the geodesic distance. On the case of computing the genuine geodesic, the distance between observations outside a particular local neighborhood can be obtained using the Floyd-Warshall algorithm [8]. In this research however, the graphical distance is considered as an approximation of the geodesic. All the distances are stored in a matrix $D_N$ which is converted to a positive definite Gram matrix using (2) The Gram matrix is then decomposed into its spectral components by the Eigen value decomposition (EVD) and projected onto the subspace of its principal values [5].

$$\mathbf{B} = -\frac{1}{2} \mathbf{H} d_M \mathbf{H}, where\ \mathbf{H} = \mathbf{I} - \frac{1}{N} \mathbf{1}\mathbf{1}^T. \tag{2}$$

Computing $D_M$ has a complexity of order $N^3$ in addition to $N^3$ order computations for the EVD [9] where N is the number of observations. This, in turn, makes the implementation of the Isomap practically unmanageable when facing high-dimensional datasets which are normally required to solve complicated processing tasks. A series of extensions have been made on most of the classical dimensionality reduction methods based on selecting a subset of landmark point out of the data in order to reduce the computational complexity of computing $D_M$ [3, 11]. Specifically, about the Isomap, it is suggested to select a set of random points throughout the space covered by the observations [7, 13]. The data graph and neighborhoods are henceforward constructed around the landmarks, *i.e.*, instead of computing all the pairwise distances and storing them in $D_N \in \mathbb{R}^{N \times N}$ with $d_{ij} = d_X(x_i, x_j)$, the distances from the landmarks to the points to their k nearest neighbors are computed and stores in $D_{kN} \in \mathbb{R}^{k \times N}$ where k indicates the number of landmark points. The short path trees in this algorithm are drawn from the landmarks to all the other data points using Dijkstra's algorithm. As a result, the complexity order of the Isomap algorithm is reduced to $KM^2$ with K the number of landmarks and M the dimensionality of the ambient space.

The random choice of landmarks may lead to some of them being either too close or too far from one another. Shi et. al. suggests a revision step of removing the initially selected landmarks which are included in the neighborhood of another landmark [13]. When defining the neighborhood as the K-nearest observations, some points with a short Euclidean distance and a large geodesic might be included in the network around an observation. This issue is referred to as a short circuit in [13] and is tackled by defining a density metric around each point-to-point path. The paths with a low density around are most likely to be short circuits. Therefore, in the Robust Landmark Isomap [13], Removes the paths which data density around them is lower than a specific threshold. These paths tend to connect the outliers to other observations.

Selection of landmarks plays a major role in proper representation of data, and thus their forthcoming interpretations. In order to overcome the improper distribution of landmarks along the data space, Manazhy suggests to apply K-means clustering and set the cluster centroids as the landmarks [5]. The centroids $\mu_i$ of the clusters $c_i$ are the solution to the optimization problem in (3).

$$\underset{c}{\mathrm{argmin}} \sum_{i=1}^{N_C} \sum_{x \in c_i} d(x, \mu_i). \qquad (3)$$

where $d(x, \mu_i)$ denotes the Euclidean distance measure.

In [10], Isomap was specifically extended for the classification tasks where each data point is represented by its geodesic distance to all the other points. The finally formed adjacency matrix is then transformed into a lower dimensional subspace so that to maximize the distinguishability between the classes. This transformation is formulated by the Fisher's linear discriminant analysis (FLDA) mapping.

Inspired by [10], Manazhy also extends the clustered landmark Isomap by concatenating an FLDA block to the aforementioned method. According to [5] the within-class and between-class variances in the clustered landmark FLDA Isomap are calculated as given in (4) and (5).

$$\boldsymbol{S}_B = \sum_{i=1}^{N_C} N_i (\boldsymbol{\mu}_i - \boldsymbol{\mu})(\boldsymbol{\mu}_i - \boldsymbol{\mu})^T, \qquad (4)$$

$$\boldsymbol{S}_W = \sum_{i=1}^{N_C} \sum_{x_k \in c_i} (x_k - \boldsymbol{\mu})(x_k - \boldsymbol{\mu})^T, \qquad (5)$$

where $\mu$ is the mean value of all the samples and $\mu_i$ is the mean values of samples in the *i*'th cluster.

Ultimately the projection matrix is given by (6).

$$\boldsymbol{W}_{FLD} = \underset{W}{\mathrm{argmax}} \frac{|W^T S_B W|}{|W^T S_W W|}. \qquad (6)$$

Solving analytically for the projection map of (6), the columns of $\boldsymbol{W}$ will correspond to the generalized eigenvectors of $\boldsymbol{S}_B$ and $\boldsymbol{S}_W$. The projection matrix $\boldsymbol{W} \in \mathbb{R}^{M \times m}$ is finally obtained by selecting the set of $m$ eigenvectors corresponding to the $m$ largest generalized eigenvalues of $\boldsymbol{S}_B$ and $\boldsymbol{S}_W$.

This closed-form solution however is hard to compute because the generalized EVD problem is of computational complexity order of $N^3$. As was initially claimed, the ambient space normally attains high dimensions which poses great difficulty on solving for generalized Eigenvectors. As in this context $S_B$ and $S_W$ tend to be of full rank with non-decaying eigenvalues, partial decompositions are not possible. Thus, the bottleneck of the generalized EVD will remain a major issue of the computation process.

### 3 Low-Rank Isomap (Proposed)

It is a general assumption over many real-world datasets that the underlying geometrical structure of the observations lie in/near a union of low-dimensional subspaces [14]. Benefited from this fact, the Low-Rank Representation (LRR) problem characterizes the basis vectors of a set of linear subspaces the union of which establishes the most similar structure to the underlying geometry of the data [15-18]. The characterized union of linear subspaces is primarily used for subspace clustering and semi-supervised classification in the literature [15-18]. However, in this paper, we discuss how this problem is equivalent to establishing a corresponding graph in a lower dimensional space. The low-dimensional graph is modeled with a low-rank coefficient matrix which holds the major information of the complete structure. Later on, we show that the partial eigenvalue decomposition of this low-rank matrix is equivalent to the complete EVD of the pairwise distance matrix D$_{KN}$ in the Isomap problem to achieve a lower computational complexity, where *K* is the number of landmarks and *N* shows the number of observations.

This paper studies the application of the LRR to concentrate the geometrical structure information in the first few generalized eigenvalues of $S_B$ so as to allow the use of partial generalized value decomposition. Solving for only the first few generalized eigenvalues reduces the computational complexity of the Landmark Extended Isomap [5], leading to the proposed Low-Rank Isomap algorithm.

#### 3.1 Low-Rank Representation

According to the LRR problem for a matrix $S = [s_1, s_2, \dots, s_N]$ with $s_i \in \mathbb{R}^M; i = 1,2,\dots,N$ can be modeled as $S = SZ + E$ where $Z \epsilon \mathbb{R}^{N \times N}$ is the coefficients matrix and $E \epsilon \mathbb{R}^{M \times N}$ is the reconstruction error matrix. If a nontrivial solution can be found for $Z$, maintaining the reconstruction error small, $S$ is then said to be self-expressive [11]. This means that every column can be reconstructed by linear combination of the other columns with a negligible error. In order to avoid the trivial solution of $Z = I$, we impose a set of constraints on the optimization problem such as low-rankness and sparsity. A variety of conditions in the specific application of image clustering is studied in [11, 19-22]. We use a basic definition of the LRR problem defined as in (7).

$$\min_{Z,E} rank(Z) + \lambda \|E\|_1 \quad s.t. \quad S = SZ + E. \tag{7}$$

As the $rank(.)$ function is not convex, we relax it by using the nuclear norm given by (8).

$$\min_{Z,E} \|Z\|_* + \lambda \|E\|_1 \quad s.t. \quad S = SZ + E. \tag{8}$$

Intuitively speaking, the low-rankness of $Z$ implies the limited number of the subspaces around the union of which the columns of $S$ are distributed as *M* dimensional points. A constraint of sparsity can also be added to the basic problem in order to encourage reconstruction of each column using only the most similar columns to it in terms of the Euclidean distance. If $Z$ is viewed as the adjacency matrix of an *N*-dimensional graph with *N* nodes, then the reconstruction of each node is performed using the closer nodes in the Euclidean space, maintaining a lower number of edges and avoiding long edges. A constraint of non-negativity will limit the reconstruction of each node based on the neighbors which form a convex set. Application of both of these constraints can lead to better results in terms of the reconstruction error, preserved information through the low-rank mapping, and the final approximated rank of $Z$ [11, 19, 20]. Therefore, the optimization problem is written as in (9).

$$\min_{Z,E} \|Z\|_* + \beta \|Z\|_1 + \lambda \|E\|_1 \quad s.t. \quad S = SZ + E, Z \geq 0. \tag{9}$$

This problem can be solved using the Adaptive Direction of Multipliers Methods (ADMM) [23]. Derivation of the recursive update formulas are presented in the following.

### 3.2 Numerical Optimization of LRR Problem

The auxiliary variable $J \in \mathbb{R}^{N \times N}$ is introduced along with an extra condition as

$$\min_{Z,J,E} \|J\|_* + \beta \|Z\|_1 + \lambda \|E\|_1 \quad s.t. \quad S = SZ + E, Z = J, J \geq 0. \tag{10}$$

The solution of (10) is equivalent to minimizing the augmented Lagrangian function given by (11), in which $M_1$ and $M_2$ are the Lagrangian Multipliers. In order to minimize the Lagrangian function, it is assumed that all variables except one are constant where the unimodal optimizations here will result in closed-form solutions. The recursive formula for updating $Z$ is given by (12).

$$\begin{aligned} \mathcal{L}(J, Z, E, M_1, M_2) \\ &= \|Z\|_* + \lambda \|J\|_1 + \beta tr(Z \times_n \mathbf{L} \times_n Z^T) + \gamma \|E\|_1 + \langle M_1, Y - YZ - E \rangle \\ &+ \langle M_2, Z - J \rangle + \frac{\mu}{2}(\|Y - YZ - E\|_F^2 + \|Z - J\|_F^2) \end{aligned} \tag{11}$$

$$Z_{k+1} = \Theta_{(\eta_1)^{-1}}(Z_k - \nabla_Z q(Z_k)/\eta_1) \tag{12}$$

Where $Z$ is updated using (14) and $\nabla_Z q(Z_k)$ is the gradient of $q(Z)$ defined by (13) and $\Theta_\varepsilon(A)$ is the Singular Value Thresholding (SVT) function of $A$ introduced by (14) [24].

$$q(Z) = \frac{\mu}{2}\left\|S - SZ - E_k + \frac{1}{\mu}M_1^k\right\|_F^2 + \frac{\mu}{2}\left\|Z - J_k + \frac{1}{\mu}M_2^k\right\|_F^2 \tag{13}$$

$$\Theta_\varepsilon(A) = US_\varepsilon(\Sigma)V^T, \tag{14}$$

Where $S_\varepsilon(x)$ is a soft thresholding function defined as in (15).

$$S_\varepsilon(x) = sgn(x)\max(|x| - \varepsilon, 0). \tag{15}$$

In [24], it is proved that the given closed-form solution is valid as far as the following condition is satisfied in (16).

$$\eta_1 > \mu(1 + \|S\|_2^2). \tag{16}$$

Updating $J$ and $E$ is then performed using (17) to (20) by using the same definition of $S_\varepsilon(x)$.

$$\min_{J \geq 0} \lambda \|J\|_1 + \frac{\mu}{2} \left\| J - (Z + \frac{1}{\mu} M_2) \right\|_F^2, \tag{17}$$

$$J_{k+1} = \max \{ S_{\frac{\lambda}{\mu}} \left( Z_{k+1} + \frac{1}{\mu} M_2^k \right), 0 \}, \tag{18}$$

$$\min_{E} \gamma \|E\|_1 + \frac{\mu}{2} \left\| E - (S - SZ + \frac{1}{\mu} M_1^k) \right\|_F^2, \tag{19}$$

$$E_{k+1} = S_{\frac{\gamma}{\mu}}(S - SZ + \frac{1}{\mu} M_1^k). \tag{20}$$

Updating the Lagrangian multiplier is performed at the end of each recursion cycle as follows [25].

$$M_1^{k+1} = M_1^k + \mu_k(S - SZ_{k+1} - E_{k+1}), \tag{21}$$

$$M_2^{k+1} = M_2^k + \mu_k(Z_{k+1} - J_{k+1}), \tag{22}$$

$$\mu_{k+1} = \min(\mu_{max}, \rho_k \mu_k), \tag{23}$$

Where $\rho_k$ is given in (24).

$$\rho_k = \begin{cases} \rho_0, if \max \{\eta_1 \|Z_{k+1} - Z_k\|, \mu_k \|J_{k+1} - J_k\|, \mu_k \|E_{k+1} - E_k\|\} \leq \varepsilon_2 \\ 1 \end{cases}. \tag{24}$$

The convergence criteria are given by the two conditions in Eq. (25) and (26).

$$\frac{\|S - SZ_{k+1} - E_{k+1}\|}{\|S\|} < \varepsilon_1 \tag{25}$$

$$\max \{\eta_1 \|Z_{k+1} - Z_k\|, \mu_k \|J_{k+1} - J_k\|, \mu_k \|E_{k+1} - E_k\|\} \leq \varepsilon_2. \tag{26}$$

All the undefined values are constant hyperparameters which must be tuned according to the application. In this work, the assigned values in the experiments are presented in Section 4.

### 3.3 Partial Generalized Eigenvalue Decomposition

The Isomap algorithm is computationally burdened by two stages of $N^3$ complexity order. The first stage is the establishment of the full graph on the data points with weighted edges. The complexity is reduced by establishing a graph based on a set of landmarks selected as the centroids of the K-means algorithm [5]. The second stage belongs to the computation of the generalized EVD in (6) with the complexity order of $N^3$, which can be sidestepped by applying a low-rank projection on the matrix $S_B$. As a result of substituting $S_B$ with its low-rank version $Z_B$, we will be able to perform partial generalized EVD in the maximization problem of (6) by solving for the projection matrix $W \in \mathbb{R}^{M \times m}$. Here, $M$ shows the dimensionality of the ambient space and $m$ is the dimensionality of the low-dimensional space, to which the data points are transferred using the Isomap algorithm.

We argue that the aim of computational complexity reduction in (6) is achieved by sole decomposition of $S_B$ due to the theorem of matrix rank under multiplication. On the other hand, the generalized EVD of $S_B$ and $S_W$ equals to the plain EVD of $S_W^\dagger S_B$ with † denoting the generalized rank preserving inverse operator [12]. By the notion of rank theorem and assuming the rank is preserved under inversion and generalized inversion [12], the generalized EVD on $S_B$ and $S_W$ can be reduced to the partial generalized EVD on the low-rank counterpart of $S_B$, namely $Z_B$ and the full-rank $S_W$. The computation must be performed by solving for a number of generalized eigenvalues equal to the reduced rank of $Z_B$. the Fast-Isomap is summarized in Alg. 1.

**Algorithm 1.** Low-Rank Isomap

Input: $X \in \mathbb{R}^{N \times M}$, $K_{max}$, $N_C$
Output: $Y \in \mathbb{R}^{N \times m}$
1: Initialize: $C = [\mu 1, \mu 2, \dots \mu NC]$
2: Optimize $C$ using K-means algorithm
3: Calculate $S_B$ and $S_W$ using (4) and (5)
4: Decompose $S_B = S_B Z_B + E_B$ using Algorithm 2
5: Perform sorted partial Generalized Eigenvalue Decomposition on low-rank $S_B$ and $S_W$ for $m$ largest generalized eigenvalues
6: Form $W = [w_1, w_2, \dots, w_d] \in \mathbb{R}^{m \times M}$ using the corresponding generalized eigenvectors
7: Project the data into the low-dimensional space using $y_i = W x_i$

**Algorithm 2.** Low-Rank Projection

Input: $S_B$
Output: $Z_B$
1: Initialize $Z = I_r, J = I_r, E = 0, M_1 = 0, M_2 = 0, \beta = 1.0, \lambda = 0.02, \mu_0 = 10^{-6}, \mu_{max} = 10^6, \rho_0 = 2.5, k = 0, \varepsilon_1 = 10^{-6}, \varepsilon_2 = 10^{-2}, \eta_1$
Repeat until convergence:
2: Update $Z_B^{k+1}$ using (12)
3: Update $J^{k+1}$ using (18)
4: Update $E^{k+1}$ using (20)
5: Update the Lagrangian multipliers $M_1^{k+1}$, $M_2^{k+1}, \mu$ using (21), (22), and (23)
6: Update $\rho_{k+1}$ using (24)
7: $k = k + 1$
8: Check convergence using (25) and (26) and stop the repetitions if converged.

A report of the computational complexity of the Alg. 1 is presented in Table 1 where $K$ is the number of the landmark points, $M$ is the dimensionality of the ambient space, and $m$ is the dimensionality of the low-rank space or the final estimated rank of $S_B$.

**Table 1.** Computational complexity of different stages in Low-Rank Isomap.

| Operation | Computational Complexity |
| --- | --- |
| K-means | $\mathcal{O}(KNM)$ |
| Calculation of $D_{MN}$ | $\mathcal{O}(KNM\log(N))$ |
| Calculation of $S_B$ and $S_W$ | $\mathcal{O}(NM^2)$ |
| Low-rank representation of $S_B$ | $\mathcal{O}(NM^2)$ |
| Partial generalized EVD of $S_B$ and $S_W$ | $\mathcal{O}(m^3)$ |

The computational complexity of stages in low-rank representation of SB is presented in Table 2. Note that $m \ll M$ and hence a reduction from $\mathcal{O}(M^2)$ to $\mathcal{O}(m^2)$ is a considerable alteration.

Table 2. Computational Complexity of the stages in Low-Rank Representation

| Operation | Computational Complexity |
|---|---|
| Update $Z_B$ | $\mathcal{O}(Nm^2)$ |
| Update $J$ | $\mathcal{O}(NM^2)$ |
| Update $E$ | $\mathcal{O}(NM^2)$ |

In conclusion, the Low-Rank Isomap algorithm will be performed in $\mathcal{O}(kNM^2)$ complexity order which for high-dimensional and large datasets is considerably smaller that the complexity order of the original Isomap. We argue that the low-rank representation of $S_B$ concentrates the information in the first few eigenvalues, resulting in a smaller final approximated dimensionality for the latent space with higher clustering accuracy therein. The performance of the proposed Low-Rank Isomap algorithm and the similar state-of-art extensions of the Isomap algorithm are presented in the following.

## 4 Experimental Results

The Low-Rank Isomap algorithm is applied to the image datasets shown in Table 3.

Table 3. Image datasets.

| Dataset | Reference | Image original size | Number of observations | Number of categories |
|---|---|---|---|---|
| AT&T | (http://www.uk.research.att.com/face-database.html/) | 112×92 | 400 | 40 |
| Yale | (http://cvc.yale.edu/) | 231×195 | 165 | 15 |
| USPS | (https://www.kaggle.com/bistaumanga/usps-dataset ) | 16×16 | 20,000 | 10 |
| ETH80 | (http://yann.lecun.com/exdb/mnist/ ) | 256×256×3 | 6,560 | 8 |

AT&T dataset provide images of human faces of 40 people under gesture variations with 10 images for each. Yale dataset contains 165 GIF images of 15 subjects under lighting and gesture variations. The USPS dataset has 7291 train and 2007 test grayscale images of handwritten digits. ETH80 contains visual object images from 8 different categories including apples, cars, cows, cups, dogs, horses, pears and tomatoes from multiple views. Regarding each dataset, the algorithm faces one of the main challenges in object recognition task as well as providing natural changes in the descriptive matrices $S_B$ and $S_W$. It is noteworthy that the images are reshaped as vectors stacked upon one another forming $X \in \mathbb{R}^{N \times M}$. All the experiments are performed on four AMD FX-8800 Radeon R7 CPUs with a clock rate of 2.10 GHz using MATLAB simulation environment.

In order to show the effect of LRR on the sorted normalized generalized eigenvalues of $S_B$ and $S_W$, they are plotted in Fig. 1 to 4 along with the normalized generalized eigenvalues without the application of the LRR. The generalized eigenvalues tend to fall reluctantly when no low-rank projections are made. But with the application of the LRR, the majority of the signal energy is concentrated in the first few eigenvalues followed by a rapid decline. This significant fall in the absolute values, yields in accurate low-rank assumption over $S_B$. For instance, the rank of original $S_B$ in AT&T dataset is approximated at 390 which is reduced to 5 after the projection. As the Yale dataset involves larger illumination variations in addition to changes in the posture, the number of considerable eigenvalues after the projection, remains larger than that in other datasets. Finally, the generalized eigenvectors corresponding to the largest eigenvalues are stacked upon the projection matrix $W$ in $y_i = W x_i$ where $x_i$ are data points in the ambient space and $y_i$ are the ones in the low-rank space.

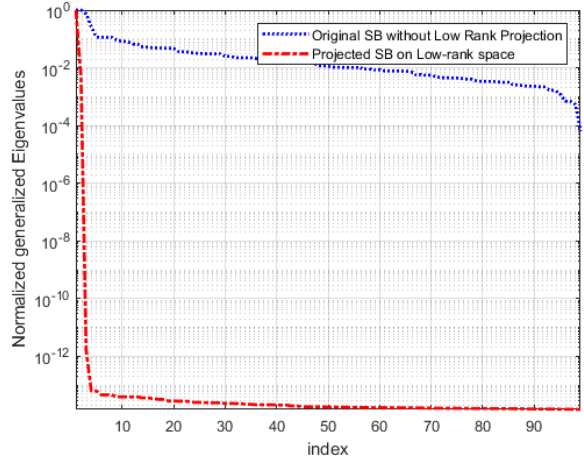

**Fig. 1** Sorted and normalized generalized eigenvalues of $S_B$ and $S_W$ in AT&T datasets with and without low-rank projection.

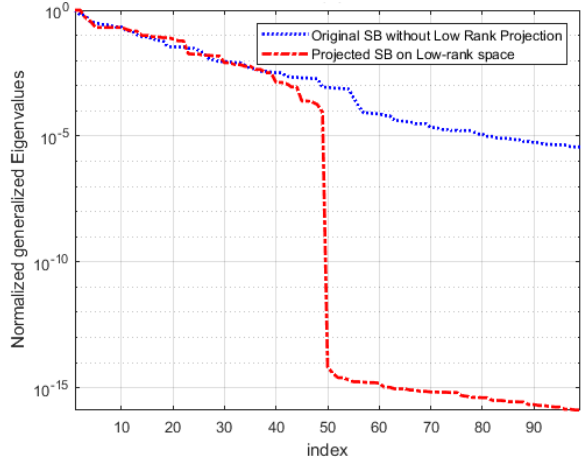

**Fig. 2** Sorted and normalized generalized eigenvalues of $S_B$ and $S_W$ in Yale datasets with and without low-rank projection.

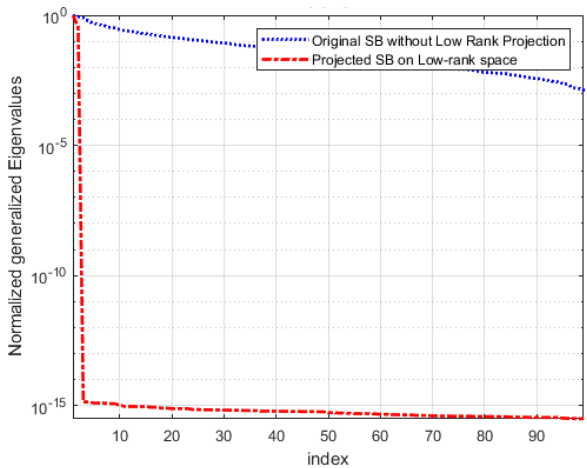

**Fig. 3** Sorted and normalized generalized eigenvalues of $S_B$ and $S_W$ in USPS datasets with and without low-rank projection.

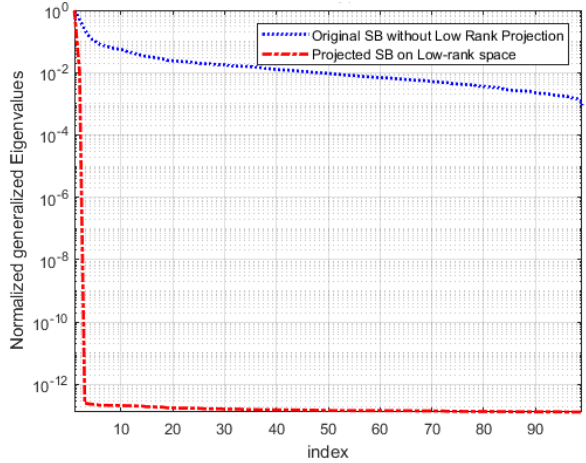

**Fig. 4** Sorted and normalized generalized eigenvalues of $S_B$ and $S_W$ in ETH-80 datasets with and without low-rank projection.

Table 4 shows the computational complexity order for the Low-Rank Isomap algorithm and several similar methods with 20 landmarks. In the former methods the majority of the computational expense is due to the generalized EVD. This burden is avoided in the Low-Rank Isomap algorithm by performing a partial EVD. Ultimately, the complexity of the Low-Rank Isomap algorithm remains noticeably smaller than the other methods when the dimensionality of the ambient space outweighs the number of observations. As N shows the number of observations, $M$ is the dimensionality of the ambient space, $K \ll N$ is the number of landmark points, and $m \ll M$ is the dimensionality of the latent space. Hence, the Low-Rank Isomap algorithm operates with noticeably a lower complexity in high-dimensional spaces. We can see that using the Low-Rank Isomap algorithm, the accuracy is not forfeited for the computational efficiency.

**Table 4.** Comparison of complexity orders in Isomap variants.

| Method | Main Stages | Complexity Order |
| --- | --- | --- |
| **Low-Rank Isomap (Proposed)** | Landmark selection, partial path calculation, low-rank projection, partial GEVD | $\mathcal{O}(KNM(1 + log(N)) + 2NM^2 + m^3)$ |
| Extended Clustered Isomap [5] | Landmark selection, partial path calculation, full GEVD | $\mathcal{O}(KNM(1 + log(N)) + NM^2 + M^3)$ |
| Robust Landmark Isomap [13] | Landmark revision, partial path calculation, full GEVD | $\mathcal{O}(NK^2 + KNMlog(N) + NM^2 + M^3)$ |
| Random Landmark Isomap [7] | Partial path calculation, full GEVD | $\mathcal{O}(KNMlog(N) + ND^2 + M^3))$ |
| Isomap [7] | Full path calculation, full GEVD | $\mathcal{O}(2N^3)$ |

### 4.2 Classification

To determine the quality of the Low-Rank Isomap algorithm in dimensionality reduction and preserving the information given in the ambient space, classification tasks are performed on the data points in the low dimensional space. The comparison is made among the Low-Rank Isomap algorithm, Extended Clustered Isomap [5], Robust Landmark Isomap [17], and Random Landmark Isomap [7] which are the state-of-art methods in the discussed setting. In order to ensure the performance is not degraded by the computational modifications, the ground level results given by the main Isomap algorithm are also presented [7].

Classification in the latent low-dimensional space is performed using the Fisher's Linear Discriminant Analysis [26] for all the aforementioned algorithms. All of the results are tested using leave-one-out cross validation. The reported accuracies show the rate of true predictions normalized to one. The leave-one-out cross validation accuracy is plotted against the number of landmarks in Figs. 5 to 8. It is observed that the proposed method stands well above the others in terms of accuracy.

Tables 5 and 6 shows the best achieved results of all the reference algorithms on all the datasets. The parameters according to the best achieved results are set as stated in Alg. 1 and 2.

Table 5. Best achieved accuracy percentages on selected datasets.

|  | AT&T | Yale | USPS | ETH |
|---|---|---|---|---|
| **Low-Rank Isomap (Proposed)** | **98.2** | **99.8** | **94.3** | **98.5** |
| *Extended Clustered Isomap [5]* | 69.8 | 87.3 | 93.1 | 88.7 |
| *Robust Landmark Isomap [13]* | 87.4 | 75.4 | 80.6 | 78.1 |
| *Random Landmark Isomap [7]* | 71.0 | 80.2 | 66.2 | 69.8 |
| *Isomap [7] (ground level)* | 82.3 | 63.1 | 71.6 | 80.9 |

Table 6. Best achieved time consumption in minutes on selected datasets.

|  | AT&T | Yale | USPS | ETH |
|---|---|---|---|---|
| **Low-Rank Isomap (Proposed)** | **61.2** | **11.6** | 32.2 | **18.7** |
| *Extended Clustered Isomap [5]* | 698.2 | 60.5 | 20.6 | 60.8 |
| *Robust Landmark Isomap [13]* | 688.3 | 102.2 | 23.5 | 57.1 |
| *Random Landmark Isomap [7]* | 654.4 | 55.9 | **16.9** | 48.8 |
| *Isomap [7] (ground level)* | 1354.1 | 203.5 | 95.7 | 235.7 |

It is observed that the Low-Rank Isomap algorithm achieves better results in a shorter time when the dimensionality of the ambient space is large. However, when the number of observations grow the computations will become heavier as well. In the context of assessing high dimensional data and noting the substantial growth of complexity in the previous methods, they literally impractical when facing high dimensional ambient spaces. Whereas the Low-Rank Isomap algorithm grows with considerably lower complexity along with the ambient space dimensionality. In all the experiments, the Low-Rank Isomap algorithm achieves higher accuracy in a shorter time as the result of concentrating all the information in a first sorted eigenvalues and applying the partial generalized EVD approach.

The number of landmarks plays a crucial role in the pairwise distances and relative locations. A certain number of landmarks are needed in order to be able to approximate the geodesic with an acceptable error. With fewer landmarks, the global structure will not be recovered at all, whereas adding redundant landmarks would cost the computational efficiency with no gain in the geodesic estimation and hence classification accuracy. In these experiments, the accuracy tends to fluctuate merely below the best achieved result when the number of landmarks is increased beyond a certain level. Figs. 5 to 8 illustrate the changes in the accuracy versus the increase in the number of landmarks for all the four datasets and the state-of-art methods.

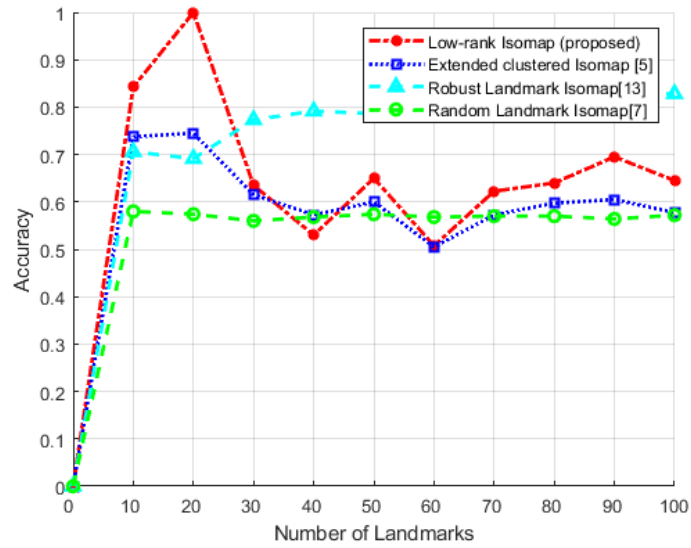

**Fig. 5** Comparison of classification (FLDA and leave-one-out cross-validation) accuracy versus the number of landmarks in AT&T datasets in Low-Rank Isomap (Proposed), Extended Clustered Isomap [5], Random Landmark Isomap [7], and Robust Landmark Isomap [13]. The dimensionality of the latent space equals to two.

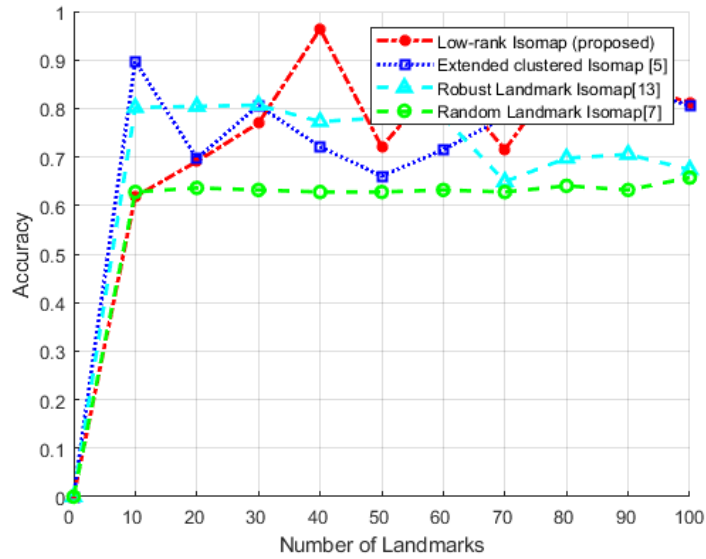

**Fig. 6** Comparison of classification (FLDA and leave-one-out cross-validation) accuracy versus the number of landmarks in Yale datasets in Low-Rank Isomap (Proposed), Extended Clustered Isomap [5], Random Landmark Isomap [7], and Robust Landmark Isomap [13]. The dimensionality of the latent space equals to two.

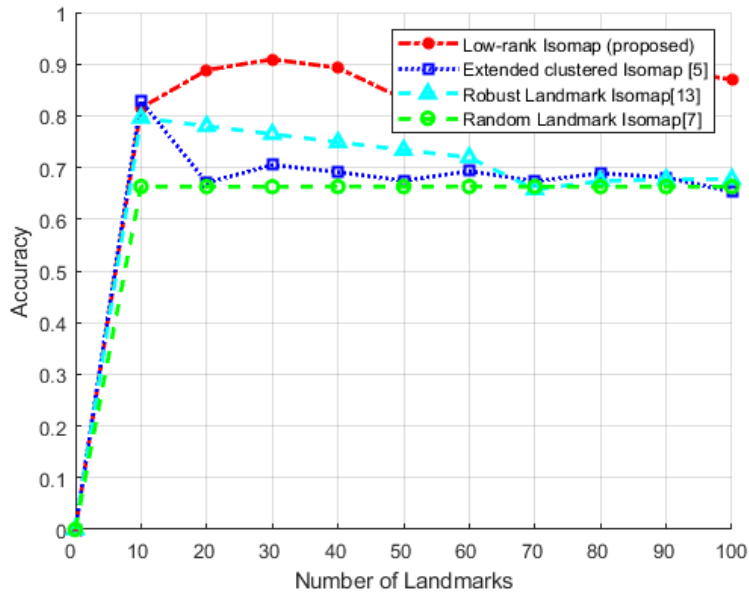

**Fig. 7** Comparison of classification (FLDA and leave-one-out cross-validation) accuracy versus the number of landmarks in USPS datasets in Low-Rank Isomap (Proposed), Extended Clustered Isomap [5], Random Landmark Isomap [7], and Robust Landmark Isomap [13]. The dimensionality of the latent space equals to two.

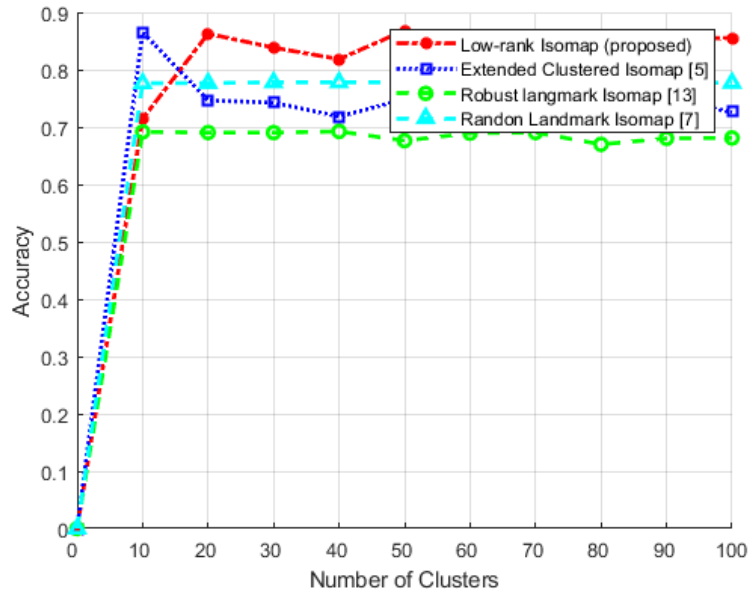

**Fig. 8** Comparison of classification (FLDA and leave-one-out cross-validation) accuracy versus the number of landmarks in ETH datasets in Low-Rank Isomap (Proposed), Extended Clustered Isomap [5], Random Landmark Isomap [7], and Robust Landmark Isomap [13]. The dimensionality of the latent space equals to two.

Figs. 9 to 12 demonstrate the performance under growing number of dimensions in the latent space. It is observed that the performance of the Low-Rank Isomap algorithm is highly maintained with varying latent space dimensionality.

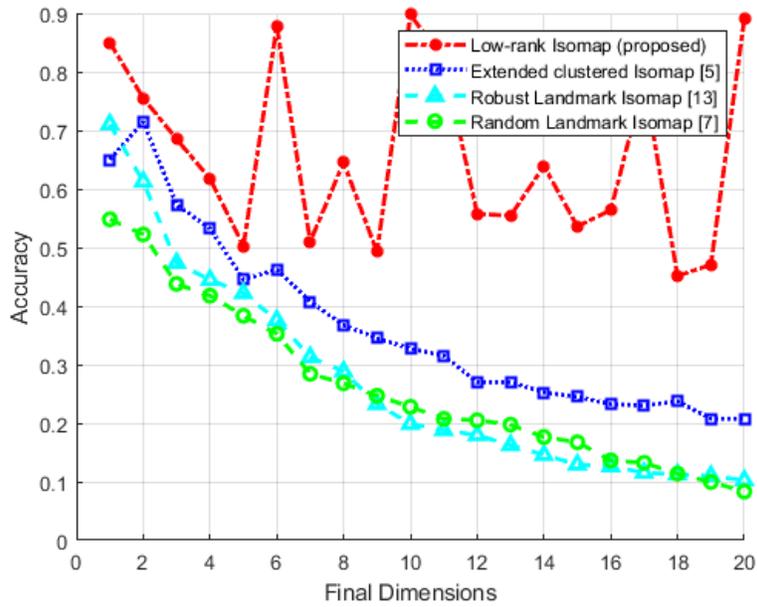

**Fig. 9** Comparison of classification (FLDA and leave-one-out cross-validation) accuracy versus the latent space dimensionality in AT&T dataset in Low-Rank Isomap (Proposed), Extended Clustered Isomap [5], Random Landmark Isomap [7], and Robust Landmark Isomap [13]. The number of landmarks equals to 20.

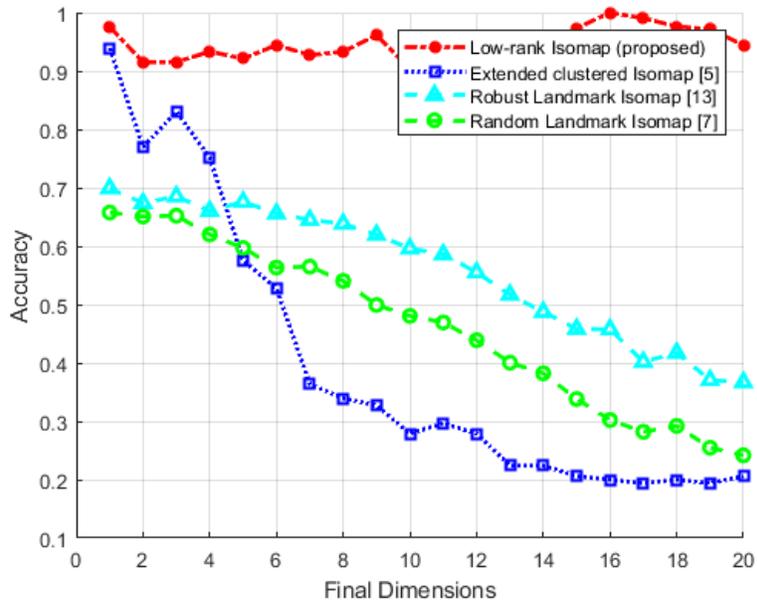

**Fig. 10** Comparison of classification (FLDA and leave-one-out cross-validation) accuracy versus the latent space dimensionality in Yale dataset in Low-Rank Isomap (Proposed), Extended Clustered Isomap [5], Random Landmark Isomap [7], and Robust Landmark Isomap [13]. The number of landmarks equals to 20.

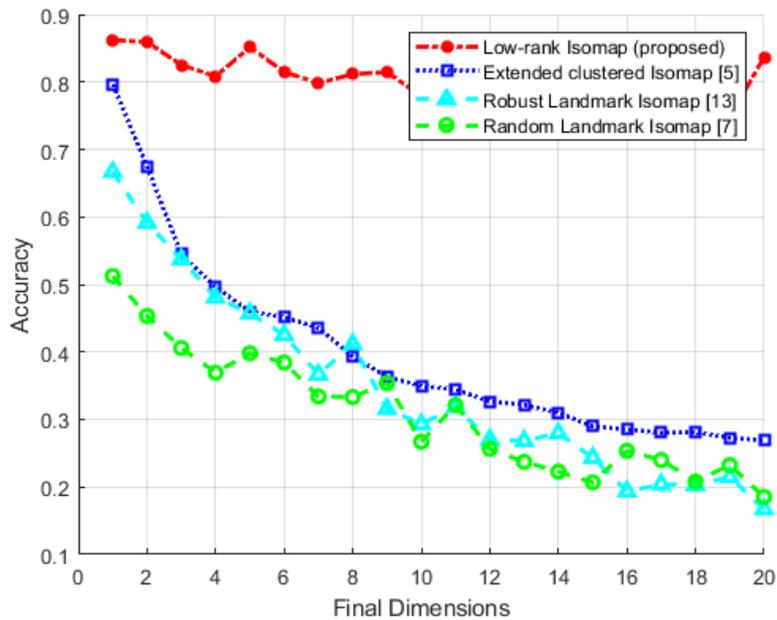

**Fig. 11** Comparison of classification (FLDA and leave-one-out cross-validation) accuracy versus the latent space dimensionality in USPS dataset in Rank Isomap (Proposed), Extended Clustered Isomap [5], Random Landmark Isomap [7], and Robust Landmark Isomap [13]. The number of landmarks equals to 20.

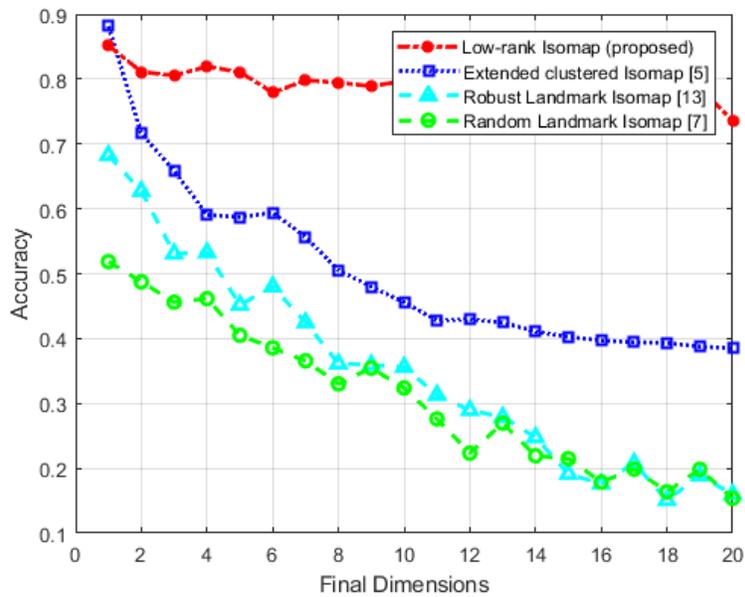

**Fig. 12** Comparison of classification (FLDA and leave-one-out cross-validation) accuracy versus the latent space dimensionality in ETH80 dataset in Low-Rank Isomap (Proposed), Extended Clustered Isomap [5], Random Landmark Isomap [7], and Robust Landmark Isomap [13]. The number of landmarks equals to 20.

When allowing the latent space dimensionality to rise, the Low-Rank Isomap algorithm preserves its performance due to the attenuation of the changes in irrelevant and redundant directions that provide no useful information for the final classification. On the contrary, the previous methods fall in performance, when the dimensionality of the latent space increases. This, in turn, blocks the application of such methods when the latent space dimensionality is not arbitrarily chosen.

## 5 Conclusion

In this paper, the practical issues of implementing Isomap as a commonly used non-linear dimensionality reduction method were discussed. We accordingly introduced the Low-Rank Isomap algorithm. In order to reduce the computational complexity of Isomap, the graphical distance, as an approximation of the geodesic, was computed based on a set of landmarks identified by the centroids of the K-mean algorithm. To enhance the distinguishability of the data points, a partial generalized EVD was performed on within-class and between-class variances after projection of the dataset onto a low-rank space. Some experiments were conducted to compare the Low-Rank Isomap algorithm with other state-of-art extensions of Isomap. Using the proposed Low-Rank Isomap algorithm, the computational complexity of Isomap was reduced from $\mathcal{O}(2N^2)$ to $\mathcal{O}(NM^2)$ ($M$ is the dimensionality of the ambient space and $N$ is the number of observations). As a result of concentrating the information along a few directions is the eigenspace, the accuracy of classification with leave-one-out cross validation was increased in all the experiments. Thus, the Low-Rank Isomap algorithm could be applied on datasets with larger numbers of observations from higher dimensional measurement spaces, maintaining a higher classification accuracy in the latent space. The best achieved results stand above 95%, which shows the superiority of the proposed method over the similar state-of-art counterparts.